\documentclass[preprint]{elsarticle}
\usepackage[margin=0.618in]{geometry}
\usepackage{amsmath,amsfonts,amssymb}
\usepackage{graphicx}
\usepackage{xcolor, float, subfig, caption}
\usepackage{array}
\newcommand{\tabincell}[2]{\begin{tabular}{@{}#1@{}}#2\end{tabular}}
\usepackage{lineno,hyperref}

\modulolinenumbers[5]
\journal{Neurocomputing}
\begin{document}
\begin{frontmatter}
\title{Comparison-Based Convolutional Neural Networks for Cervical Cell/Clumps Detection in the Limited Data Scenario}

\author{Yixiong Liang}
\ead{yxliang@csu.edu.cn}
\author{Zhihong Tang}
\ead{zhihongtang@csu.edu.cn}
\author{Meng Yan}
\ead{bryant@csu.edu.cn}
\author{Jialin Chen}
\author{Qing Liu}
\author{Yao Xiang\corref{mycorrespondingauthor}}
\cortext[mycorrespondingauthor]{Corresponding author.}
\ead{yao.xiang@csu.edu.cn}
\address{School of Computer Science and Engineering, Central South University, Changsha 410083, China}

\begin{abstract}
Automated detection of cervical cancer cells or cell clumps has the potential to significantly reduce error rate and increase productivity in cervical cancer screening. However, most traditional methods rely on the success of accurate cell segmentation and discriminative hand-crafted features extraction. Recently there are emerging deep learning-based methods which train convolutional neural networks (CNN) to classify image patches, but they are computationally expensive. In this paper we propose an efficient CNN-based object detection methods for cervical cancer cells/clumps detection. Specifically, we utilize the state-of-the-art two-stage object detection method, the Faster-RCNN with Feature Pyramid Network (FPN) as the baseline and propose a novel comparison detector to deal with the limited data problem. The key idea is that classify the proposals by comparing with the reference samples of each category in object detection. In addition, we propose to learn the reference samples of the background from data instead of manually choosing them by some heuristic rules. Experimental results show that the proposed Comparison Detector yields significant improvement on the small dataset, achieving a mean Average Precision (mAP) of 26.3\% and an Average Recall (AR) of 35.7\%, both improving about \textbf{20} points compared to the baseline. Moreover, Comparison Detector improved AR by \textbf{4.6} points and achieved marginally better performance in terms of mAP compared with baseline model when training on the medium dataset. Our method is promising for the development of automation-assisted
 cervical cancer screening systems. Code is available at
 \url{https://github.com/kuku-sichuan/ComparisonDetector}.
\end{abstract}

\end{frontmatter}

\section{Introduction}
Cervical cytology is the most common and effective screening method for cervical cancer and premalignant cervical lesions \cite{davey2006effect}, which is performed by a visual examination of cytopathological analysis under the microscope of the collected cells that have been smeared on a glass slide and stained and finally giving a diagnosis report according to the descriptive diagnosis method of the Bethesda system (TBS)\cite{nayar2015bethesda}. Currently in developed countries, it has been widely used and has significantly reduced the number of deaths caused by related diseases, but it is still unavailable for population-wide screening in the developing countries \cite{saslow2012american}, partly due to the fact that it is labor-intensive, time-consuming and expensive \cite{bengtsson2014screening}. In addition, it is subjective and therefore has motivated lots of automated methods for the automation of cervical screening based on the image analysis techniques.

Over the past 30 years extensive research has attempted to develop automation-assisted screening methods \cite{zhang2014automation,phoulady2016automatic}. 
Most of them try to classify a single cell into various stages of carcinoma, which often consists three steps: cell (cytoplasm and nuclei) segmentation, feature extraction and classification. The performance of these methods, however, heavily depends on the accuracy of the segmentation and the effectiveness of the hand-crafted features.

With the overwhelming success in a broad range of applications such as image classification \cite{krizhevsky2012imagenet,he2016deep}, semantic segmentation \cite{long2015fully}, object detection \cite{ren2015faster, lin2017feature} and medical imaging analysis \cite{litjens2017survey,esteva2017dermatologist}, CNN has also been applied to the segmentation and classification of cervical cell \cite{tareef2017optimizing,zhang2017combining,lu2017evaluation,zhang2017deeppap,jith2018deepcerv}. The majority of them (e.g. \cite{tareef2017optimizing,zhang2017combining}) are trying to take advantage of CNN to improve the segmentation accuracy of cytoplasm and nuclei, but they do not provide the needed segmentation accuracy \cite{lu2017evaluation,jith2018deepcerv}, whereas once the segmentation error are taken into account, the classification accuracy would drop \cite{zhang2017deeppap}. To avoid the dependence on accurate segmentation, the patch-based methods (e.g. \cite{jith2018deepcerv}) try to use CNN to classify the image patches. However, the extraction of such patches still requires the segmentation of nuclei. The recent work \cite{zhang2017deeppap} also adopts the patch-based strategy but during the inference the random-view aggregation and multiple crop testing are needed to produce the final prediction results and thereby is time-consuming.

In this paper, we propose an efficient strategy to apply CNN for cervical cancer screening, \emph{without} any pre-segmentation step. Specifically, we exploit the contemporary CNN-based object detection methods \cite{ren2015faster,lin2017feature} to detect the cervical cytological abnormalities directly. It is straightforward and has been successfully applied for other medical image analysis \cite{liang2018end,liang2018object}, but very few works try to apply CNN-based object detection for automated cervical cytology. We attribute this to the lack of the right cervical cancer microscopic image dataset for the detection task. CNN-based object detection methods often need sufficient annotated data to obtain good generalization, but for cervical cytological abnormalities detection, collecting the large amounts of data with careful and accurate annotation is difficult partially due to the limitation by laws, the scarcity of positive samples and especially the unanimous agreement between cytopathologists \cite{stoler2001interobserver}.

To alleviate the limited data problem, we propose the named \emph{Comparison Detector}, which migrate the idea of \emph{comparison} in one/few-shot learning for image classification \cite{koch2015siamese, vinyals2016matching, snell2017prototypical, yang2018learning} into CNN-based object detection, for cervical cancer detection. Specifically, we choose the state-of-the-art object detection method, Faster R-CNN \cite{ren2015faster} with FPN \cite{lin2017feature}, as our baseline model and replace the original parameter classifier with a non-parametric one based on the idea of comparison with the prototype representations of each category, which is generated from reference samples. Furthermore, instead of manually choosing the reference images of the background category by some heuristic rules, we propose to learn them from the data. We also investigate several important factors including the generation of prototype representations of each category and the design of head model for cervical cell or cell clumps detection. Our algorithm directly operates on the whole image rather than the extracted patches based on the nuclei and hereby only need one forward propagation for each image, making the inference very efficient. In addition, the proposed method is \emph{flexible} to be integrated into other proposal-based methods.

We collect a small dataset $D_s$ and a medium dataset $D_f$ which are directly dedicated to cervical cell/clumps detection, on which we evaluate the performance of the proposed Comparison detector. When the model is learned from the small dataset $D_s$, the performance of our method is significantly better than the baseline model, i.e. Comparison detector has a mAP 26.3\% and an AR 35.7\% but the baseline model only gains a mAP 6.6\% and an AR 12.9\%. When the model is learned from the medium dataset $D_f$, our Comparison detector achieves performance with a mAP of 45.9\% compared to 45.2\%, and improves nearly 5 points comparing to baseline model with AR.

We summarize our contributions as follows: 1) We propose an end-to-end object detection method called Comparison detector to deal with the limited data problem in cervical cell/clumps detection; 2) We propose a strategy to directly learn the prototype representations of background and 3) Our method performs much better than the baseline on both small and medium dataset and has the potential applications to the real automation-assisted cervical cancer screening systems.
\begin{figure*}[t]
\centering
\includegraphics[width=\textwidth]{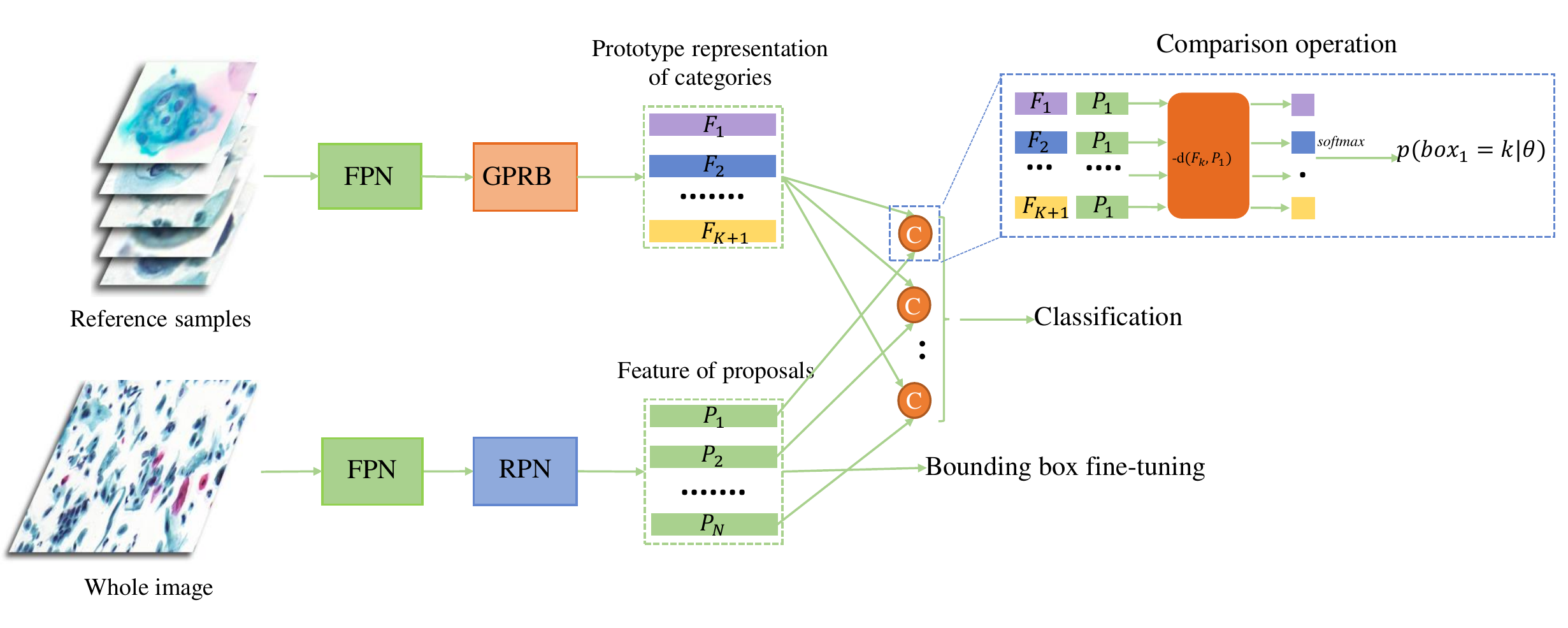}
\caption{\label{fig:1} The overall structure of Comparison detector. First, features of $n \times K$ reference samples are obtained by ResNet50 with FPN, where $K$ is number of category and $n$ is number of each category. Then, it generates the prototype representations for each category by the \textbf{Generating Prototype Representation Block (GPRB)}. At the same time, it obtains the features of proposals from the whole image through ResNet50 with FPN and RPN. It should be noted that the ResNet50 with FPN is shared. Finally, by comparing the features of each proposal with all prototype representations, we can get the category of this proposal. Only the feature of proposals are used to fine-tune the bounding box.}
\end{figure*}

\begin{figure*}[t]
\begin{minipage}{\textwidth}
\centering
\subfloat[]{
\includegraphics[width=\textwidth]{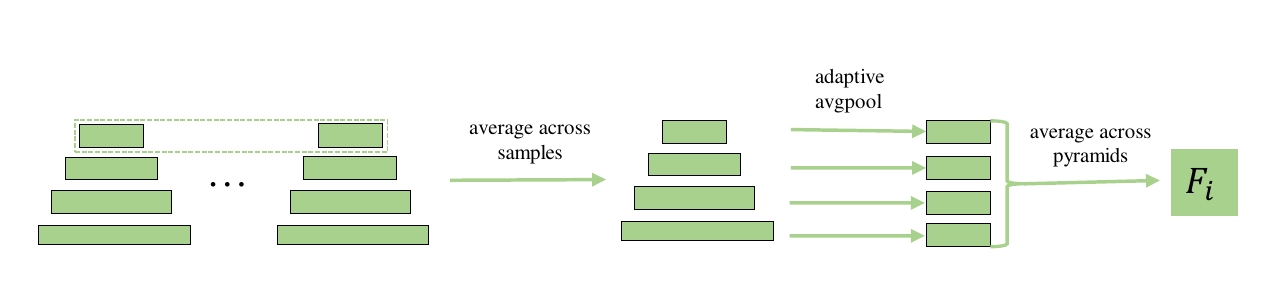}
}
\end{minipage}

\begin{minipage}{\textwidth}
\centering
\subfloat[]{
\rotatebox{90}{\includegraphics[width=0.3\textwidth]{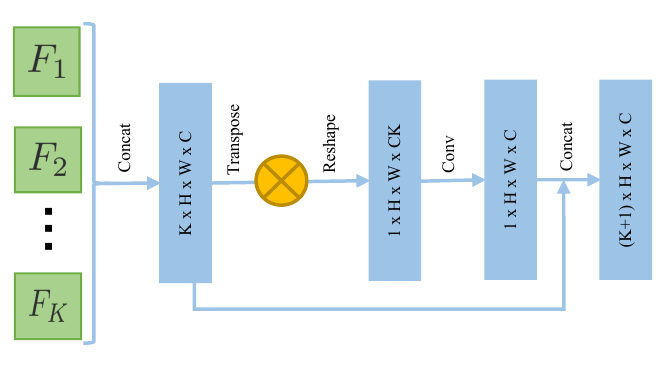}}
}
\end{minipage}
\caption{
\textbf{Generating Prototype Representation Block (GPRB)}.$K$ is number of categories exclude background. $W, H$ and $C$ are the width, height and channels of prototype representation. We called the K categories as foreground category. (a): The process of generating one foreground prototype representation from pyramid features of reference samples. (b): The operation of generating background prototype representation through foreground prototype representations and concatenating with K foreground prototype representations.}
\label{fig:2}
\end{figure*}

\section{Related work}
 \subsection{cell segmentation and classification} Traditional cytological criteria for classifying cervical cell abnormalities are based on the changes in nucleus to cytoplasm ratio, nuclear size, irregularity of nuclear shape and membrane, therefore there are numerous works focusing on the segmentation of cell or cell components (nuclei, cytoplasm) \cite{zhang2014segmentation,zhang2017graph,
 lee2016segmentation}. Although significant progress has been achieved recently, the segmentation of cell or cell components remains an open problem due to the large shape and appearance variation between cells, the poor contrast
of cytoplasm boundaries and the overlap between cells \cite{zhang2017deeppap,lee2016segmentation,lu2017evaluation}.

On the other hand, cervical cell classification methods try to differentiate the dysplastic cells from the norm cells and classify them into various stages of carcinoma. According to TBS rules \cite{nayar2015bethesda}, a large number of hand-crafted features are designed to describe the shape, texture and appearance characteristics of the nucleus and cytoplasm \cite{marinakis2009pap,phoulady2016automatic}. The resulting features are often further organized by feature selection or dimension reduction and then are fed into various classifiers (e.g. random forests, SVM, softmax regression, neural network, etc.) to perform the final classification. However, as mentioned above, the extraction of those engineered features depends on the accurate segmentation of cell or cell components. Furthermore, it is also limited by the current understanding of cervical cytology \cite{zhang2017deeppap}. To reduce the dependency on the accurate segmentation, the CNN are used to learn the features from data recently \cite{jith2018deepcerv}, but an approximate segmentation or region of interest (ROI) detection is still necessary. Although the DeepPap \cite{zhang2017deeppap} is claimed totally segmentation-free, it still needs the nucleus centroid information for training and the random-view aggregation and multiple crops testing during the inference stage, which are very time-consuming.

There are a handful public available microscopic image datasets dedicated to cervical cell segmentation such as ISBI-14\footnote{\url{https://cs.adelaide.edu.au/~carneiro/isbi14_challenge/index.html}}, ISBI-15 \footnote{\url{https://cs.adelaide.edu.au/~zhi/isbi15_challenge/index.html}}, but to our best knowledge for cervical cell classification the only public available microscopic image dataset is the Herlev benchmark dataset \cite{marinakis2009pap}, which consists of 917 single cell images corresponding to four categories of abnormal cell with different severity (namely light dysplastic, moderate dysplastic, severe dysplastic and carcinoma in situ) and three categories of normal cells (normal columnar, normal intermediate and normal superficial). The limited annotated data prevents the applications of traditional object detection methods such as Viola-Jones detector \cite{viola2004robust} or contemporary CNN-based detectors \cite{liu2018deep} to cervical cancer screening.

\subsection{CNN-based object detection} The Overfeat \cite{sermanet2013overfeat} made the earliest efforts to apply CNN for object detection and has achieved a significant improvement of more than 50\% mAP when compared to the best methods at that time which were based on the hand-crafted features. Since then, a lot of CNN-based methods \cite{chu2018deep, ren2015faster,girshick2014rich,
girshick2015fast,he2017mask,redmon2016you,redmon2017yolo9000,zhang2018single} have been proposed for high-quality object detection, which can be roughly classified into two categories: object proposal-based and proposal-free. The road-map of proposal-based methods starts from the notable R-CNN \cite{girshick2014rich} and is improved by Fast-RCNN \cite{girshick2015fast} in an end-to-end manner and by Faster R-CNN \cite{ren2015faster} to quickly generate object regions, which has motivated a lot of follow-up improvements \cite{lin2017feature,he2017mask} in terms of accuracy and speed. The proposal-free methods \cite{redmon2016you,redmon2017yolo9000} directly predict the bounding boxes without the proposal generation step. Generally, the proposal-free methods are conceptually simpler and much faster than the proposal-based methods, but the detection accuracy is usually behind that of the proposal-based methods \cite{zhang2018single}. Here we choose the Faster R-CNN \cite{ren2015faster} with FPN \cite{lin2017feature} as our baseline model but our method is compatible with other proposal-based methods.

\subsection{One/few-shot learning} One/few-shot learning is a task of learning from just one or a few training samples per class and has been extensively discussed in the context of image recognition and classification \cite{koch2015siamese,vinyals2016matching,snell2017prototypical}. Recently significant progress has been made for one/few-shot learning tackled by meta-learning or learning-to-learn strategy, which can be roughly divided into three categories: metric-based, memory-based and optimization-based. The metric-based methods \cite{koch2015siamese,vinyals2016matching,snell2017prototypical,yang2018learning} learn to compare the query image with support set images. The memory-based methods \cite{santoro2016meta} exploited the memory-augmented neural network to
quickly store and retrieve sufficient information for each classification task, while the optimization-based methods \cite{ravi2016optimization,finn2017model} aim to learning a base-model which can be fine-tuned quickly for a new classification task. All these works only tackle image classification tasks.

\subsection{Object detection with limited data} Most prior works on object detection with limited labels use semi-/weakly-supervised methods or few-example learning \cite{dong2018few} to make use of abundant unlabeled data, whereas in limited data regime there are few works focus on using few-shot learning to address this problem \cite{schwartz2018repmet,kang2019few}. Kang et al. \cite{kang2019few} decomposes the training into base-model learning and meta-model learning and train a meta-model to reweight the features extracted by the base-model to assist novel object detection. However, the training of base model still needs abundant annotated data for base classes. RepMet \cite{schwartz2018repmet} introduces a metric learning-based sub-network architecture to learn the embedding space and distribution of the training categories without using external data. However, RepMat involves an alternating optimization between the external class distribution module learning and net parameters updating, whereas our solution is a clean, single-step training framework.

\begin{figure}[t]
\centering
\begin{minipage}{0.33\textwidth}
\subfloat[]{
\includegraphics[width=0.7\textwidth]{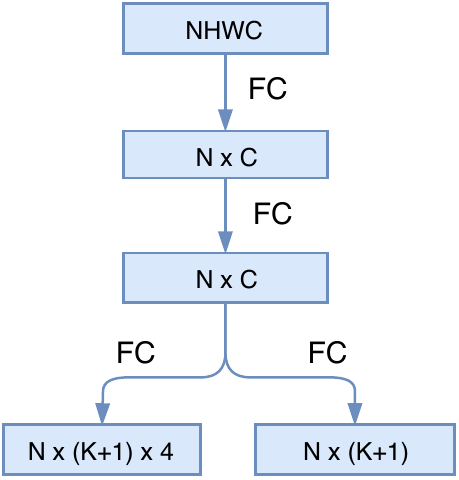}
}
\end{minipage}%
\begin{minipage}{0.33\textwidth}
\subfloat[]{
\includegraphics[width=0.7\textwidth]{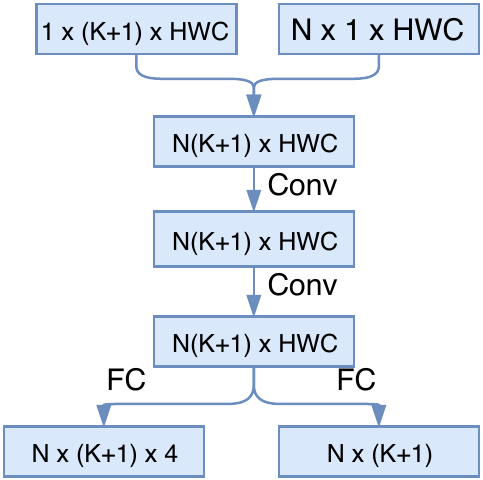}
}
\end{minipage}%
\begin{minipage}{0.33\textwidth}
\subfloat[]{
\includegraphics[width=0.7\textwidth]{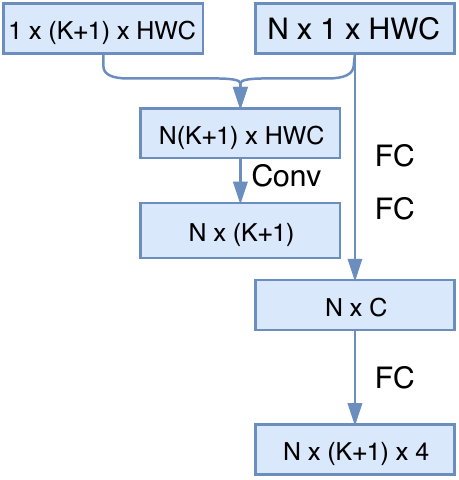}
}
\end{minipage}
\caption{ \textbf{The head for classification and regression.}
\textbf{(a)}: The head of baseline model. \textbf{(b)}: The share module in our experiments. \textbf{(c)}: The independent module in the Comparison Detector.}
\label{fig:3}
\end{figure}

\begin{figure*}[!t]
\centering
\begin{minipage}{0.5\textwidth}
\subfloat[]{
\includegraphics[width=6cm, height=6cm]{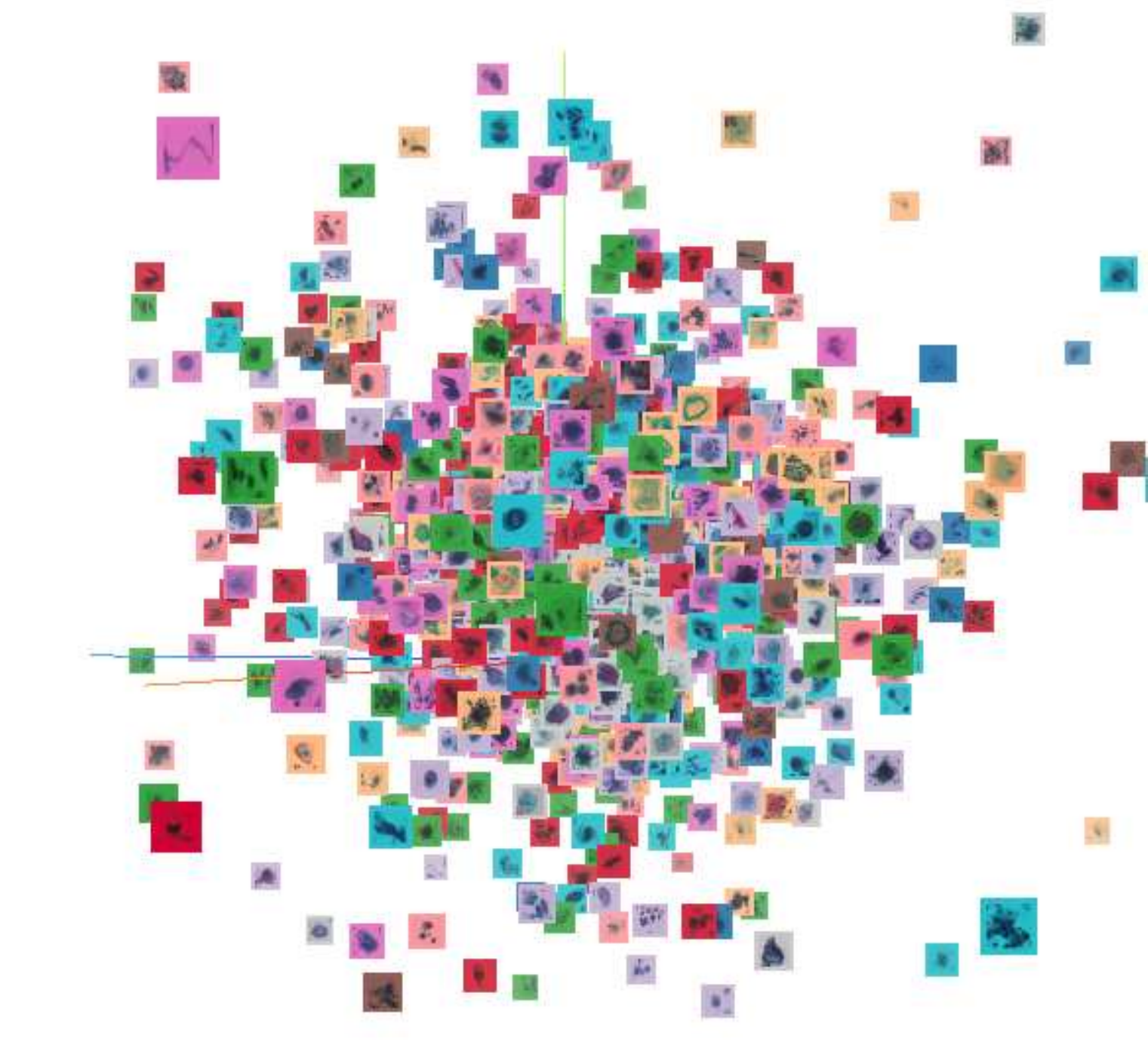}
}
\end{minipage}%
\begin{minipage}{0.5\textwidth}
\subfloat[]{
\includegraphics[width=6cm, height=6cm]{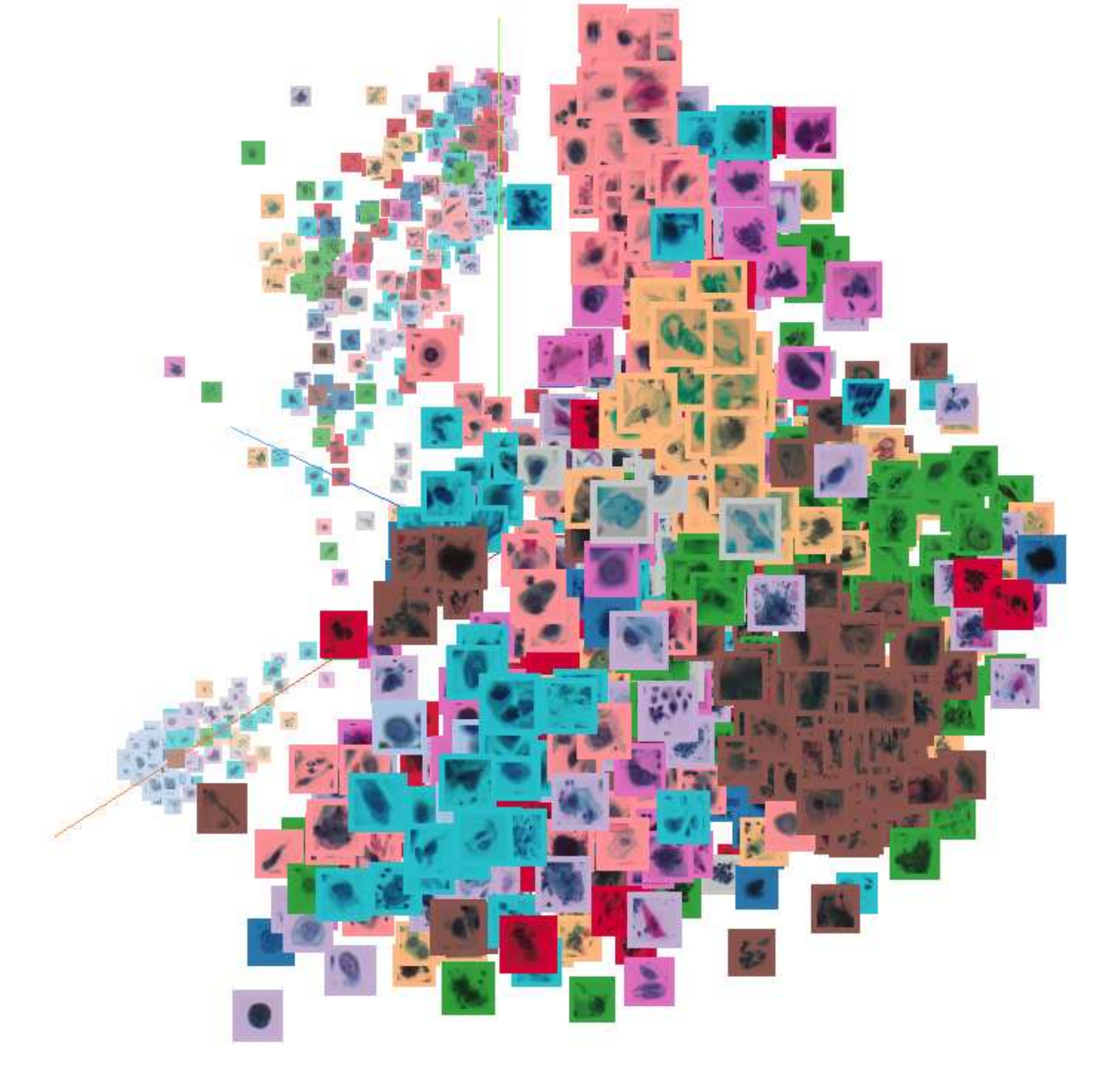}
}
\end{minipage}
\caption{\label{fig:4}
\textbf{t-SNE} visualization for reference samples. \textbf{(a)}:Visualization before learning. \textbf{(b)}:Visualization after learning.
}
\end{figure*}

\section{Comparison Detector}\label{section3}
\subsection{Basic Architecture}
Our proposed Comparison detector is based on proposal-based detection framework 
consisting of a backbone network for feature extraction, a region proposal network (RPN) for generating proposals and a head for 
the proposal classification and bounding box regression. Here we choose the Faster R-CNN with FPN \cite{lin2017feature} as our baseline. 
Then we decouple the regression and classification in the head and replace the original parameter classifier with our comparison classifier. 
Our comparison classifier introduces an inductive bias, i.e. the within-class distance is less than the between-class in the features space, 
into the model and henceforth reduces the complexity of the model and mitigates the generalization issue with small datasets to some extent \cite{battaglia2018relational}.

The framework of our Comparison detector is depicted in Fig.\ref{fig:1}, which is divided into three stages to describe. 
At the first stage, Comparison detector generates features for the reference samples and the whole images. As shown in Fig. \ref{fig:1}, 
both the features of them are computed by backbone network with FPN \cite{lin2017feature}, without adding any extra models to encode the reference samples. Assuming there are $n$ reference samples for $K$ category with $L$ levels pyramid feature. 
Let $F_{kj}^l$ be the $j$-th sample with $k$-th categories' prototype representation of the $l$-level pyramid features, 
which can be computed by average operation as follows
\begin{equation}
F_{kj}^l = F^l (R_{kj}) \label{equ:1}
\end{equation}
where $F^l(\cdot)$ and $R_{kj}$ denote the $l$-th level feature extraction function and the $j$-th reference sample of class $k$, respectively. 
At the same time, the feature $P_m$ of the $m$-th object proposal $x_m$ is generated by
\begin{equation}
P_m = F^l(x_m). \label{equ:2}
\end{equation}
It should be pointed out that categories in the training and test set are
 same in our settings unlike one/few-shot learning.

The second stage is to generate the prototype representations of each category from the reference samples' pyramid features as shown in Fig. \ref{fig:2}(a). The function of $S(\cdot)$ is to compute the final prototype representation $F_k$ for class $k$
\begin{equation}
F_{k}^l = \frac{1}{n}\sum_{j=1}^n F_{kj}^l \label{equ:3}
\end{equation}

\begin{equation}
F_k = S (\{F_k^l\}) \label{equ:4}
\end{equation}

The third stage is the head model for classification and bounding box regression (Fig. \ref{fig:3}(c)), 
consisting of a few convolutional ($Conv$) and fully connected ($FC$) layers. 
Let $d(F_k,P_m)$ be a metric function to compute the distance between the $m$-th proposal feature
($P_m$) and prototype representation of the $k$-th category ($F_k$). We will discuss this function in section \ref{sect:4.5}. Each proposal's posterior probability $p_k$ and bounding box regression $b_k$ can be obtained by
\begin{equation}
 p_k = \frac{e^{-d(F_k,P_m)}}{\sum_ie^{-d(F_i,P_m)}} \label{equ:5}
\end{equation}

\begin{equation}
b_k = b(F_k, P_m) \label{equ:6}
\end{equation}
where $b(\cdot,\cdot)$ denotes the bounding box regression function. We denote Eq. \ref{equ:5} as comparison classifier. 
The rest of the model is the same as Faster R-CNN with FPN model  \cite{lin2017feature}.

Finally, The objective function of Comparison detector is to minimize the total loss consisting of the RPN classification loss, RPN bounding box loss, 
head classification loss, head bounding box loss:
\begin{equation}
  \min_w L_{rpnc} + L_{rpnb} + \lambda L_{headc} + L_{headb} \label{equ:7}
\end{equation}
$w$ is the weight of Comparison detector. All classification losses are cross-entropy loss and bounding box losses are $\ell_1$-smooth loss \cite{girshick2015fast}.

\subsection{Generating Prototype Representation Block (GPRB)}

\subsubsection{Generating prototype representations of categories}
As shown in Eq. \ref{equ:3}, we use $n$ reference samples for each category to generate pyramid features, 
then obtains the prototype representation like Eq. \ref{equ:4}.  We'll show a possible choice for $S(\cdot)$. 
For simplicity, we directly resize the each pyramid features which is generated by reference 
samples to a fixed size, and then calculate prototype representation by averaging operation, i.e.
\begin{equation} \label{equ:8}
S(\{F_k^l\}) = \frac{1}{L}\sum_{l}r(F_k^l, s)
\end{equation}
where $r(\cdot, \cdot)$ is bilinear interpolation function 
and $s$ is the size of final features.
\subsubsection{Learning the prototype representations of background}
There are many negative proposals generated by RPN,
 so the R-CNN \cite{girshick2014rich} adds a background category to represent them. 
 In our Comparison detector, we need to select a number of reference samples for each category. Because of the overwhelming diversity,
 selecting reference samples of background is very difficult. We notice that a background region is considered to the proposal indicating 
 that it has certain similarity with categories. Therefore, it can be inferred that its prototype representations are a combination of different
 categories in the most case. So we propose to learn its prototype representations from other categories' prototype representations. First, we transpose the channels and reshape the tensor. Then, we use a simple $1\times1$ convolution operation to generate the prototype representations of background; finally, we concat all prototype representations together, as shown in Fig. \ref{fig:2}(b).

\subsection{The head for classification and regression}\label{section:3.3}
As shown in  Fig. \ref{fig:3}(a), the structure of the baseline model's head is to transform the proposal 
feature firstly and then one branch is used for classification, and another is used to predict the offset 
of the bounding box. For our Comparison detector, due to the introduction of the reference samples, 
we need to re-organise the head. The are two choices according to whether sharing the features between
bounding box regression and classification. One is that bounding box regression branch and classification branch are not shared,
as shown in Fig. \ref{fig:3}(c). Unlike the baseline model, the classifier and bounding box regressor in the head of 
 Comparison detector are independent (independent module). And the bounding box regressor only uses the features of proposal to predict the offset of the 
 bounding box. It is equivalent to
\begin{align*}
   d(F_i,P_m) & = FC(m(F_i,P_m)), \\
   b(F_i,P_m) &  = FC(FC(FC(P_m))),
\end{align*}
where $m(F_i, P_m) = Conv_3(Conv_1(|F_i - P_m|^2))$\footnote{The subscript denotes kernel size of convolution}. 
Another choice is to share features for both classification and regression, as shown in Fig. \ref{fig:3}(b), which means
\[
 b(F_i,P_m) = FC(m(F_i,P_m)).
\]
We call this method as shared module.

\begin{table*}
\renewcommand{\arraystretch}{1.3}
\caption{All experiments train on $D_f$, and the performance is evaluated on the test set. At the same time, the comparison classifier of all models directly adopts the $\ell_2$-distance. The reference samples are the same, produced by fixed mode before the experiment.}
\label{tab:1}
\centering
\begin{tabular}{|c|c|c|c|c|c|c|c|}
\hline
\rule[-1ex]{0pt}{2.5ex}\bfseries model & \bfseries \tabincell{c}{learning \\ background} & \bfseries independent mode & \bfseries \tabincell{c}{using all \\ pyramid features}  & \bfseries \tabincell{c}{twice bounding box\\ regression}& \bfseries mAP & \bfseries AR \\
\hline\hline

\rule[-1ex]{0pt}{2.5ex}  \texttt{A} &        &$\surd$ &$\surd$ &$\surd$ &31.4 &49.3\\
\hline
\rule[-1ex]{0pt}{2.5ex} \texttt{B} &$\surd$ &$\surd$ &$\surd$ &$\surd$ &34.1 &53.3\\

\hline
\rule[-1ex]{0pt}{2.5ex} \texttt{C} &$\surd$ &$\surd$ &        &$\surd$ &32.7 &50.8\\
\hline
\rule[-1ex]{0pt}{2.5ex} \texttt{D} &$\surd$ &        &$\surd$ &$\surd$ & 41.0 &51.3\\
\hline
\rule[-1ex]{0pt}{2.5ex} \texttt{E} &$\surd$ &        &        &$\surd$ & 38.9 &49.8\\
\hline
\rule[-1ex]{0pt}{2.5ex} \texttt{F} &$\surd$ &$\surd$ &$\surd$ &        & 37.7 &51.1\\
\hline
\end{tabular}
\end{table*}

\subsection{Strategies for selecting reference samples}
In our Comparison detector, we need to choose the reference samples for each category. 
A intuitive way is to select them according to the Bethesda atlas \cite{nayar2015bethesda}. 
However, there are very significant difference between the given atlas and our data due to 
the variations of the preparation and digitization of slide. Hence we resort to other feasible data-driven alternatives. 
We randomly select about 150 instances of each category from the training sets. 
The shortest side of these instances is greater than 16 pixels. Therefore we get a total of 1,560 instances as candidate reference samples 
from training sets. we can select suitable instances in these objects as our reference samples.

There are two possible ways. The first is to randomly choose several instances of each category as the reference samples. 
The second is to first map all 1,560 objects into the feature space through the ImageNet pre-trained model \cite{he2016deep} to 
get the features of each object and then use t-SNE \cite{maaten2008visualizing} for feature dimension reduction (Fig. \ref{fig:4}). 
Based on the results of t-SNE, we empirically obtain the number of clusters each category. Then we use it as the parameter for K-means. Finally, based on the result of K-means, we choose the instances which are the closest to the center of clusters as reference samples.

\begin{figure*}[t]
  \centering
  \includegraphics[width=.7\textwidth]{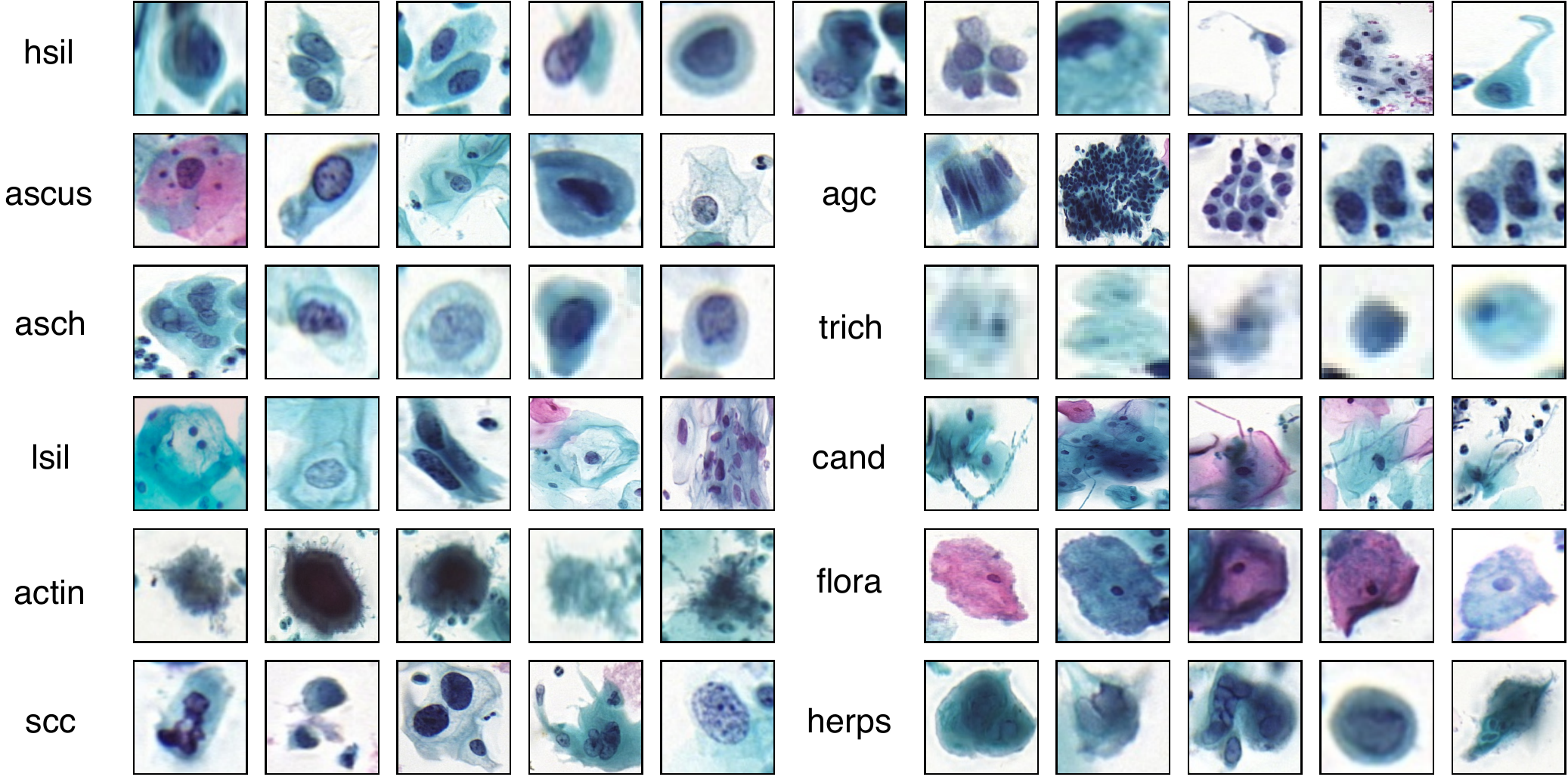}
  \caption{Object samples of entities seen in cervical cytology.}\label{fig:5}
\end{figure*}

\begin{figure*}[!t]
\centering
\includegraphics[scale=0.45]{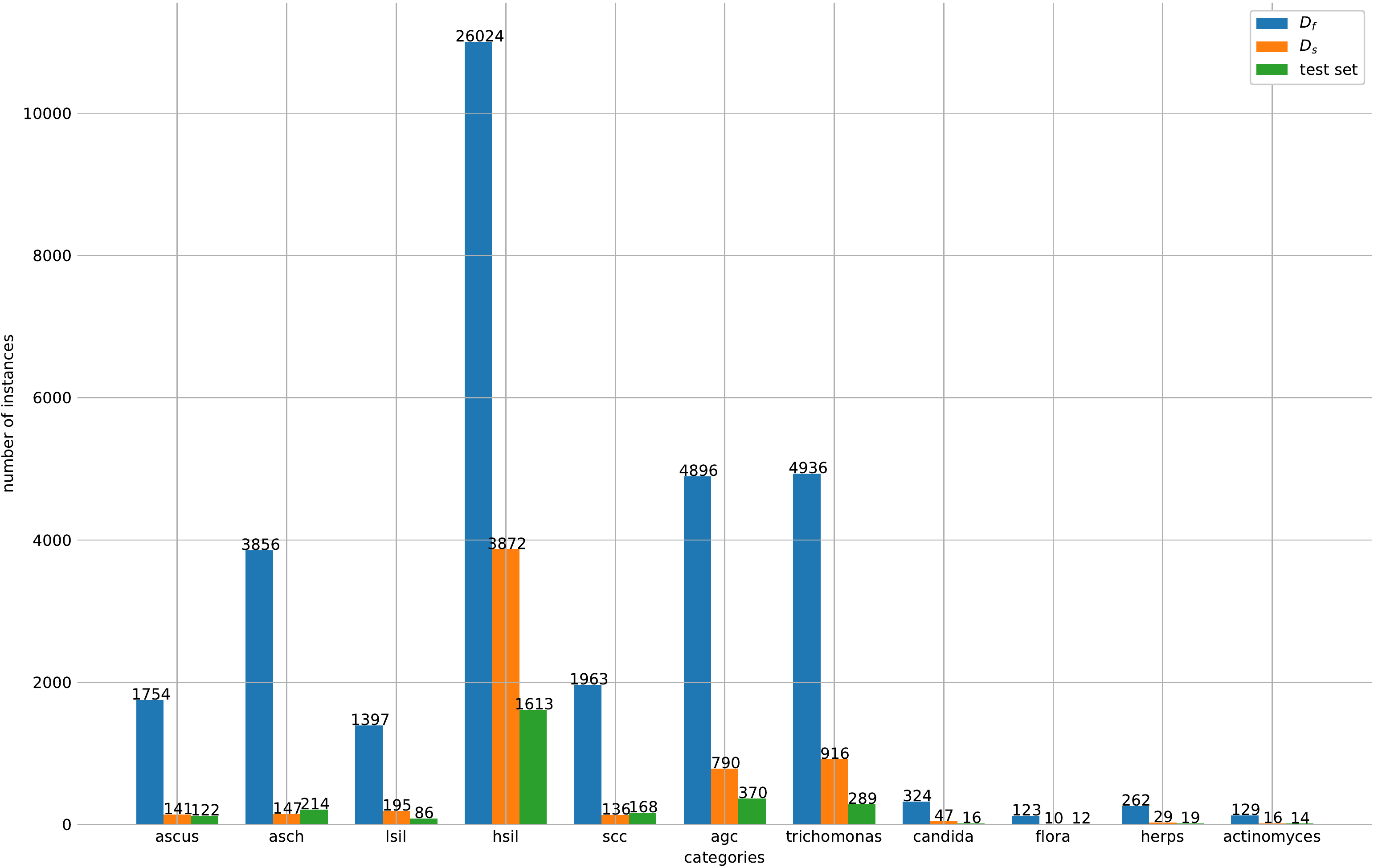}
\caption{\label{fig:6} The distribution of categories on different datasets.}
\end{figure*}

\section{Experiment and Result}

\begin{table}[t]
\caption{Model with balanced bounding box regression loss and classification loss. The metric in bracket is AR.}
\centering
\label{tab:2}
\begin{tabular}{|c|c|c|c|}
\hline
\bfseries model & \texttt{B} & \texttt{D} & \texttt{F}\\
\hline \hline
wo/blancing & 34.1(53.3) & 41.0(51.3) & 37.7(51.1)\\
\hline
w/blancing &\textbf{43.7(60.7)} & \textbf{43.5(58.9)} & \textbf{38.8(52.3)}\\
\hline
\end{tabular}
\end{table}

\begin{table*}[t]
\caption{Different comparison classifier. The numbers in the brackets denote the result after balancing the loss.}
\label{tab:3}
\centering
\begin{tabular}{|c|c|c|c|}
\hline
\bfseries comparator & \bfseries $\ell_2$-distance &\bfseries  parameterized $\ell_2$-distance  & \bfseries concat \\
\hline\hline
   mAP       &34.1(43.7)&    38.2(\textbf{44.5})           &     40.7(42.5)   \\
\hline
   AR          &53.3(60.7)&    56.8(\textbf{61.6})          &     49.1(58.1)   \\
\hline
\end{tabular}
\end{table*}

\begin{table*}[!t]
\renewcommand{\arraystretch}{1.3}
\caption{Number of clusters for each category.}
\label{tab:4}
\centering
\begin{tabular}{|c|c|c|c|c|c|c|c|c|c|c|c|}
\hline
\bfseries category & \bfseries ascus &\bfseries asch &\bfseries lsil &\bfseries hsil &\bfseries scc & \bfseries agc & \bfseries trich & \bfseries cand & \bfseries flora & \bfseries herps&\bfseries actin\\
\hline\hline
number of clusters & 3 & 4 & 2 & 4 & 2 & 3 &1 & 2 & 2 & 4 & 1 \\
\hline
\end{tabular}
\end{table*}

\subsection{Materials and experiments} \label{sect:4.1}
There are no public available benchmarks for cervical cell object detection in the community, we first establish a database consisting of 7,086 cervical microscopical images which are cropped from the whole slide images (WSIs) obtained by Pannoramic MIDI II digital slide scanner. The corresponding specimens are prepared by Thinprep methods stained with Papanicolaou stain.  Conforming to TBS categories \cite{nayar2015bethesda}, 48,587 object instance bounding boxes were labeled by experienced pathologists which belong to 11 categories, namely ASC-US (ascus), ASC-H (asch), low-grade squamous intraepithelial lesion (lsil), high-grade squamous intraepithelial lesion (hsil), squamous-cell carcinoma (scc), atypical glandular cells (agc), trichomonas (trich), candida (cand), flora, herps, actinomyces (actin). Figure \ref{fig:5} shows some examples of each category in our database. Then we randomly divide the dataset into training set $D_f$ which contains 6,667 images, test set which contains 419 images for experiment. To verify the performance of Comparison detector on the small dataset, we randomly choose 762 images from the training dataset to form a small dataset of $D_s$. The distribution of categories in each dataset is shown in the Fig. \ref{fig:6}.

In all experiments, we used ResNet50 as backbone network with ImageNet pre-trained model. 
For reference samples, we re-scale them such that their side is $w=h=224$ which is coincident with pre-trained model. 
The initial learning rate is 0.001, and then decreased by a factor of 10 at 35-th and 50-th epoch. 
Training is stopped after 60 epochs and the other parameters are the same as FPN \cite{lin2017feature}. The minibatch size is 2 images in 2 GPUs. The weight decay is 0.0001 and the momentum is 0.9.
The experiment is firstly trained on the $D_f$ to evaluate the performance of Comparison detector. 
In our setting, the reference samples are fixed in each training iteration for the stability of the training model. 
And test stage is the same.

\begin{figure*}[t]
\centering
\caption{\textbf{Comparison detector} results on the test set. These results are based on ResNet-50 with Feature Pyramid Networks, achieving a mAP of 45.9\% and AR of 63.5\%.}
\label{fig:7}
\begin{minipage}{0.24\textwidth}
\includegraphics[height=3.2cm, width=4.3cm]{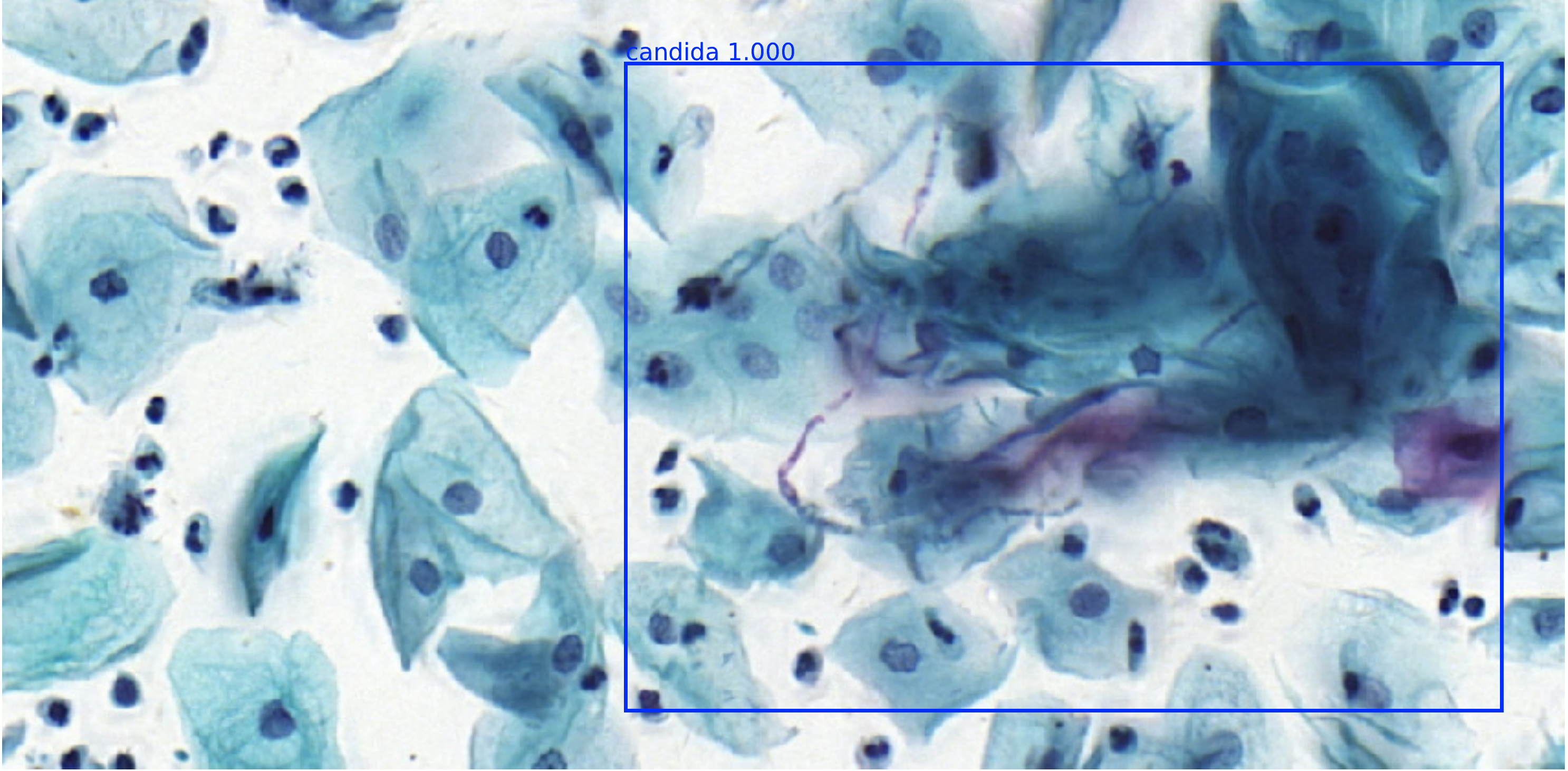}
\end{minipage}%
\begin{minipage}{0.24\textwidth}
\includegraphics[height=3.2cm, width=4.3cm]{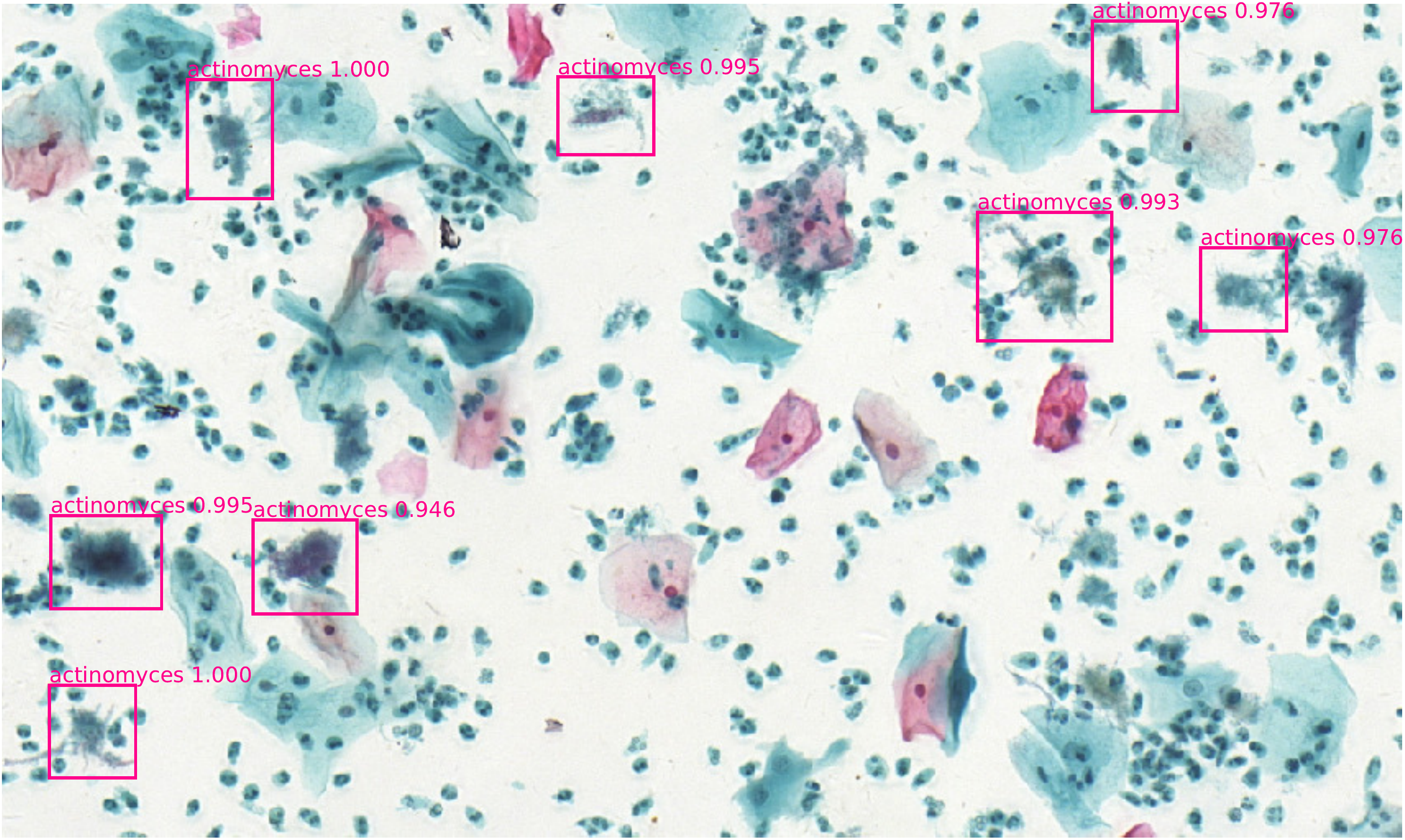}
\end{minipage}%
\begin{minipage}{0.24\textwidth}
\includegraphics[height=3.2cm, width=4.3cm]{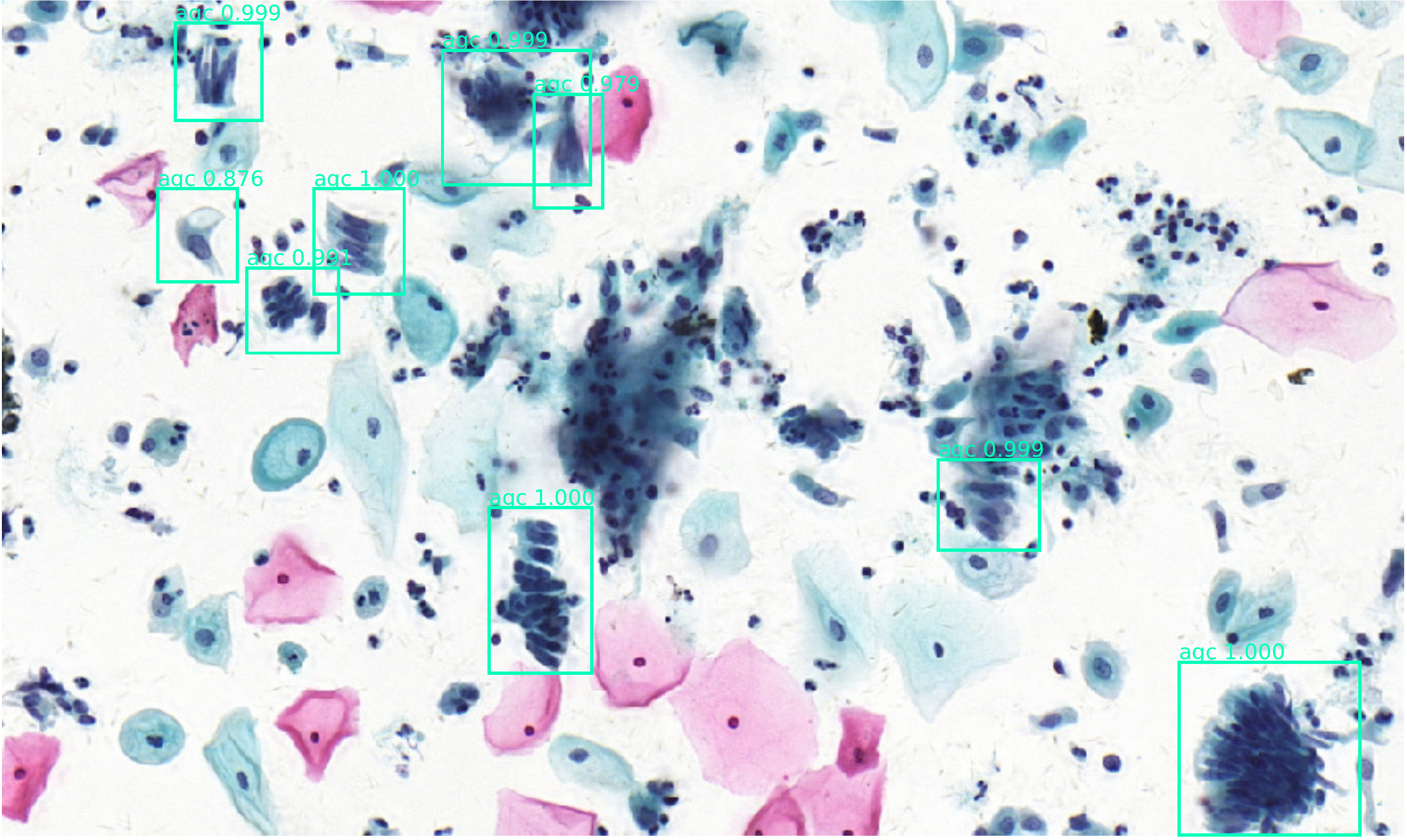}
\end{minipage}%
\begin{minipage}{0.24\textwidth}
\includegraphics[height=3.2cm, width=4.3cm]{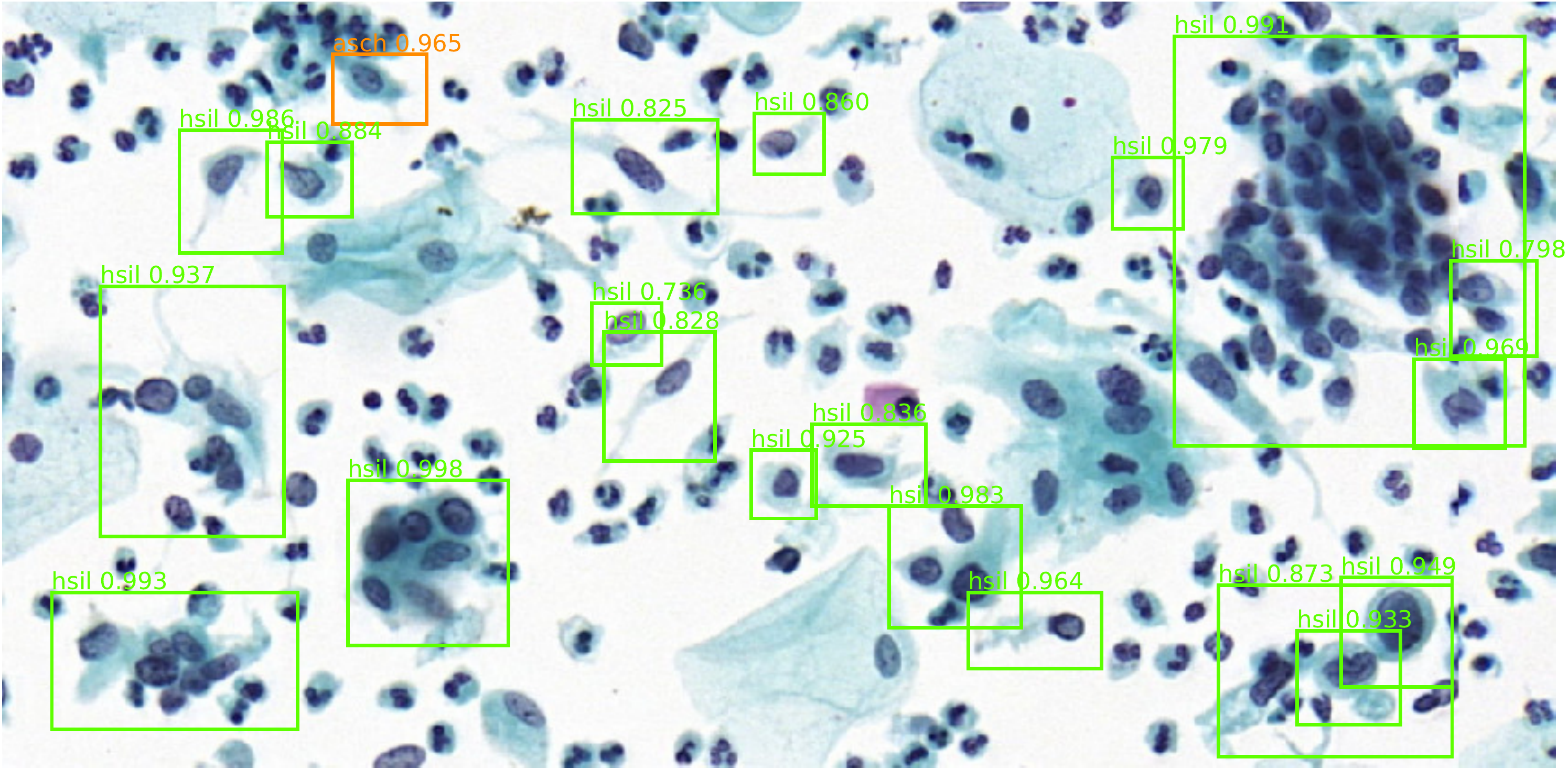}
\end{minipage}

\begin{minipage}{0.24\textwidth}
\includegraphics[height=3.2cm, width=4.3cm]{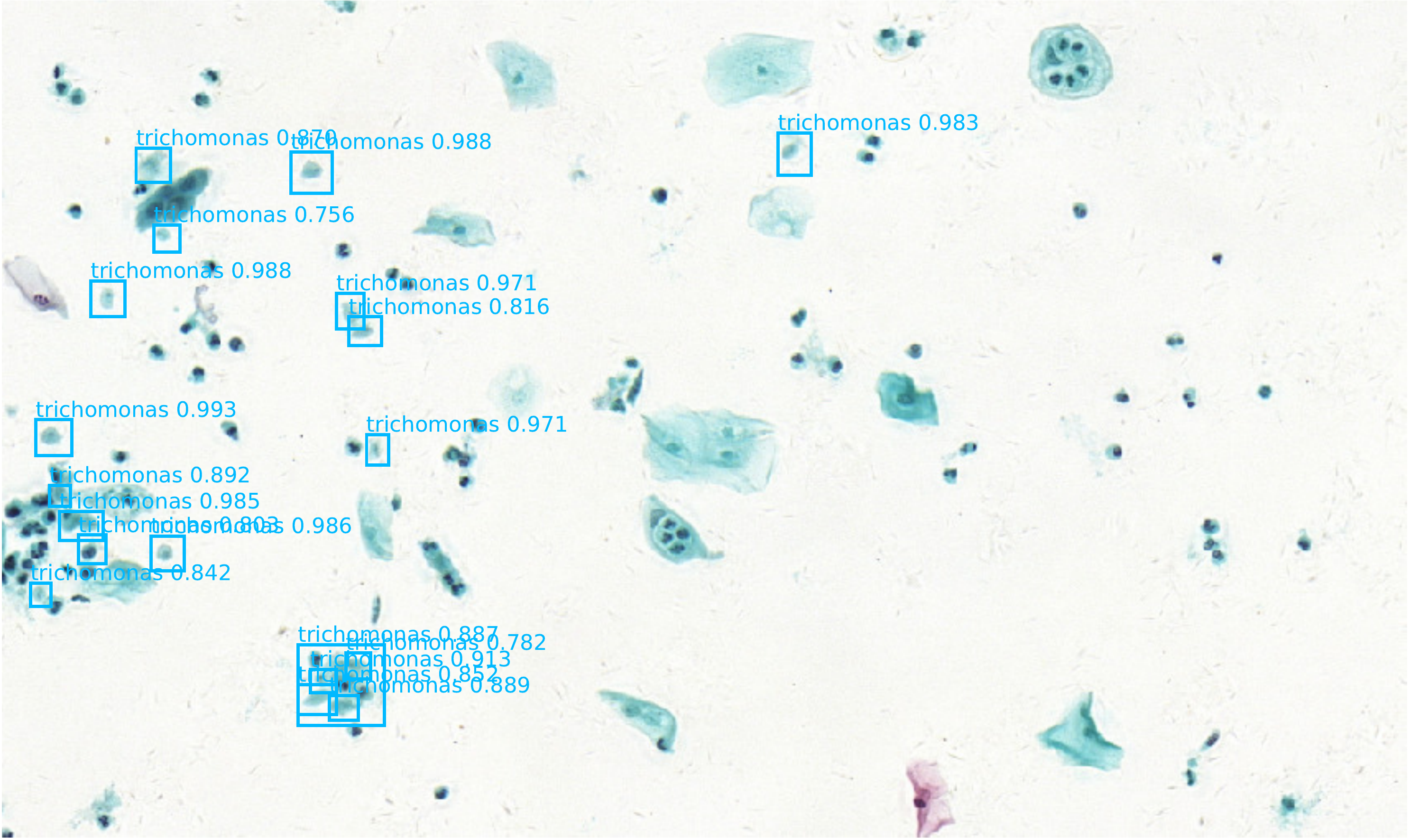}
\end{minipage}%
\begin{minipage}{0.24\textwidth}
\includegraphics[height=3.2cm, width=4.3cm]{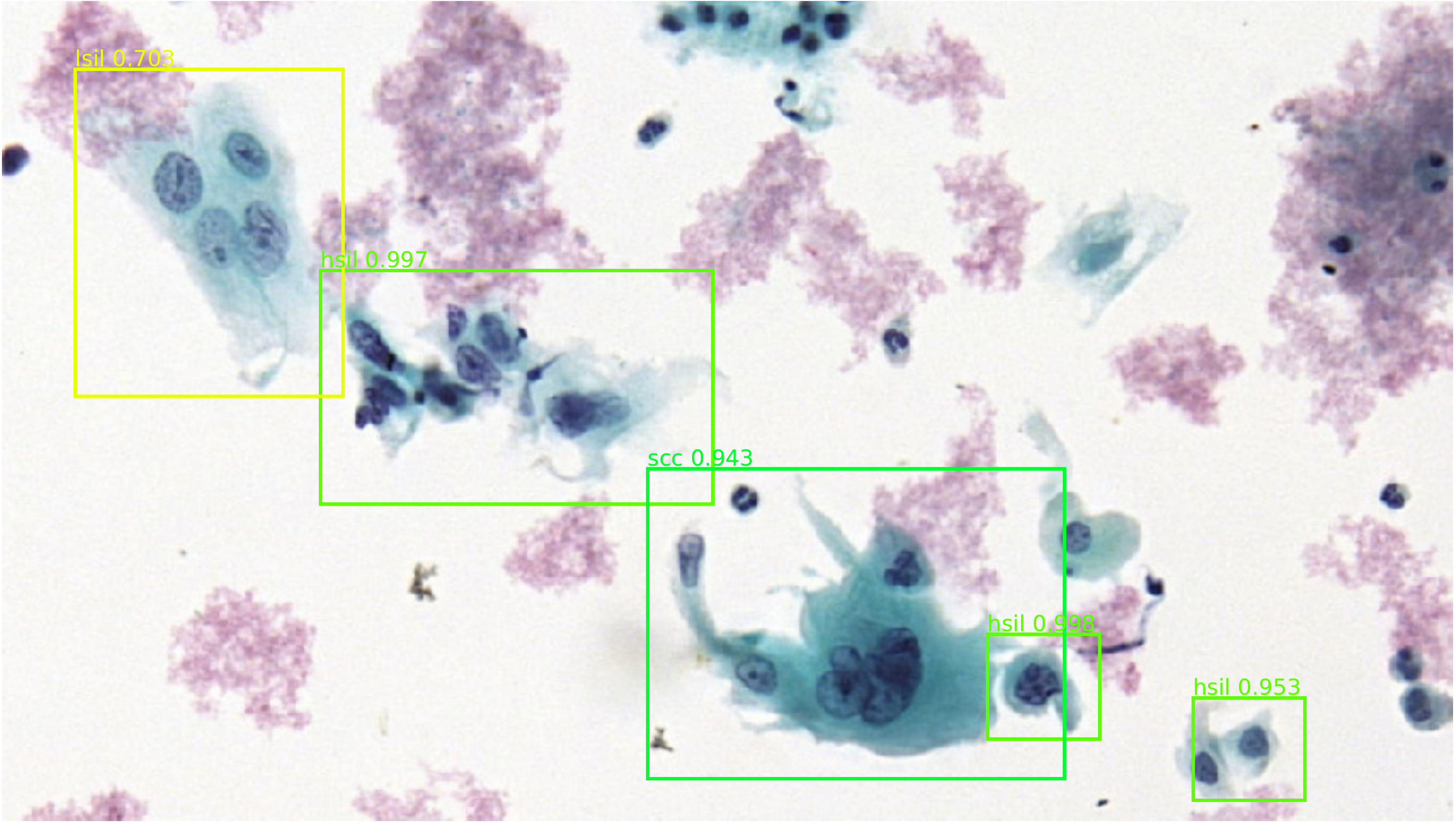}
\end{minipage}%
\begin{minipage}{0.24\textwidth}
\includegraphics[height=3.2cm, width=4.3cm]{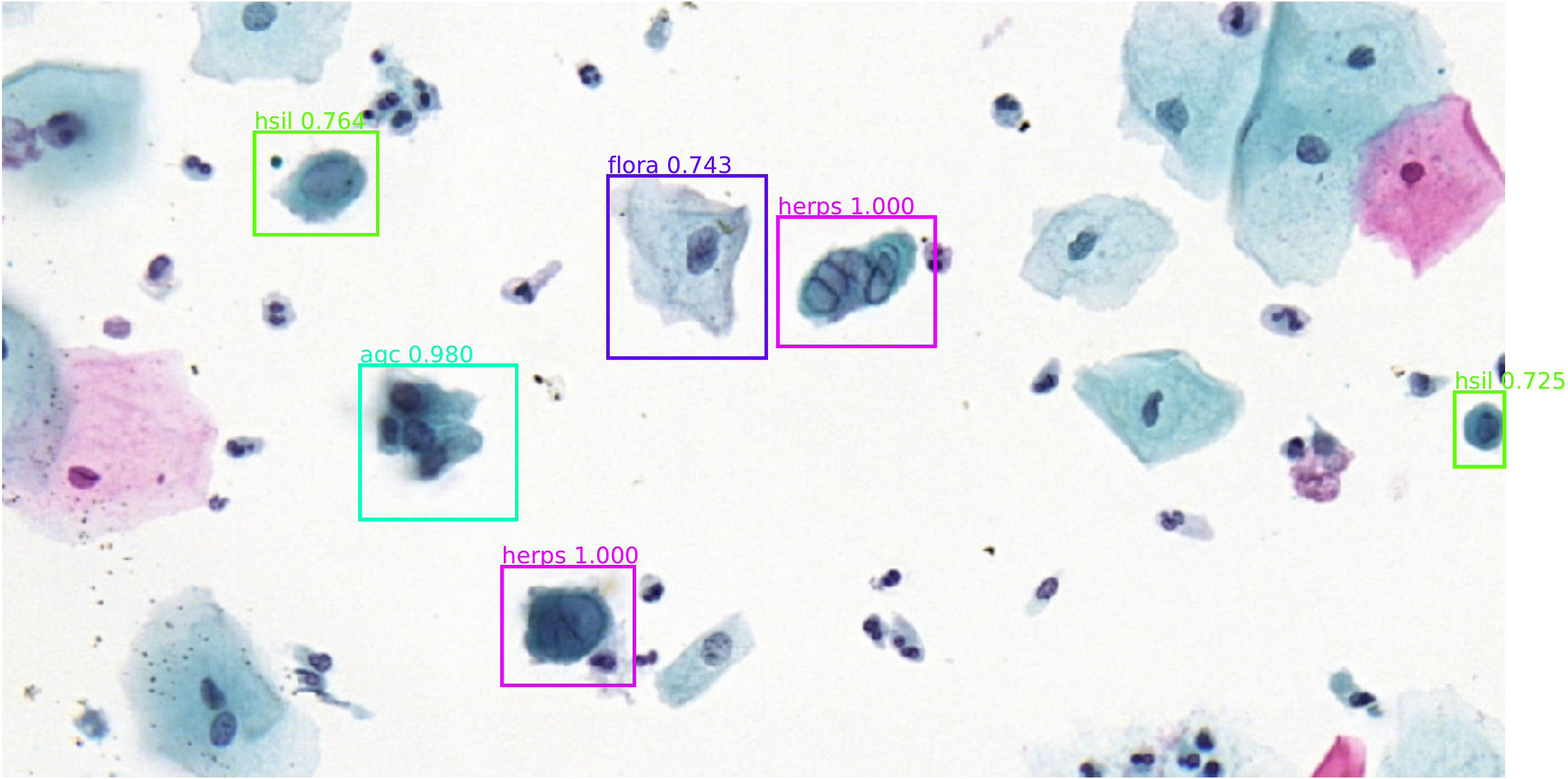}
\end{minipage}%
\begin{minipage}{0.24\textwidth}
\includegraphics[height=3.2cm, width=4.3cm]{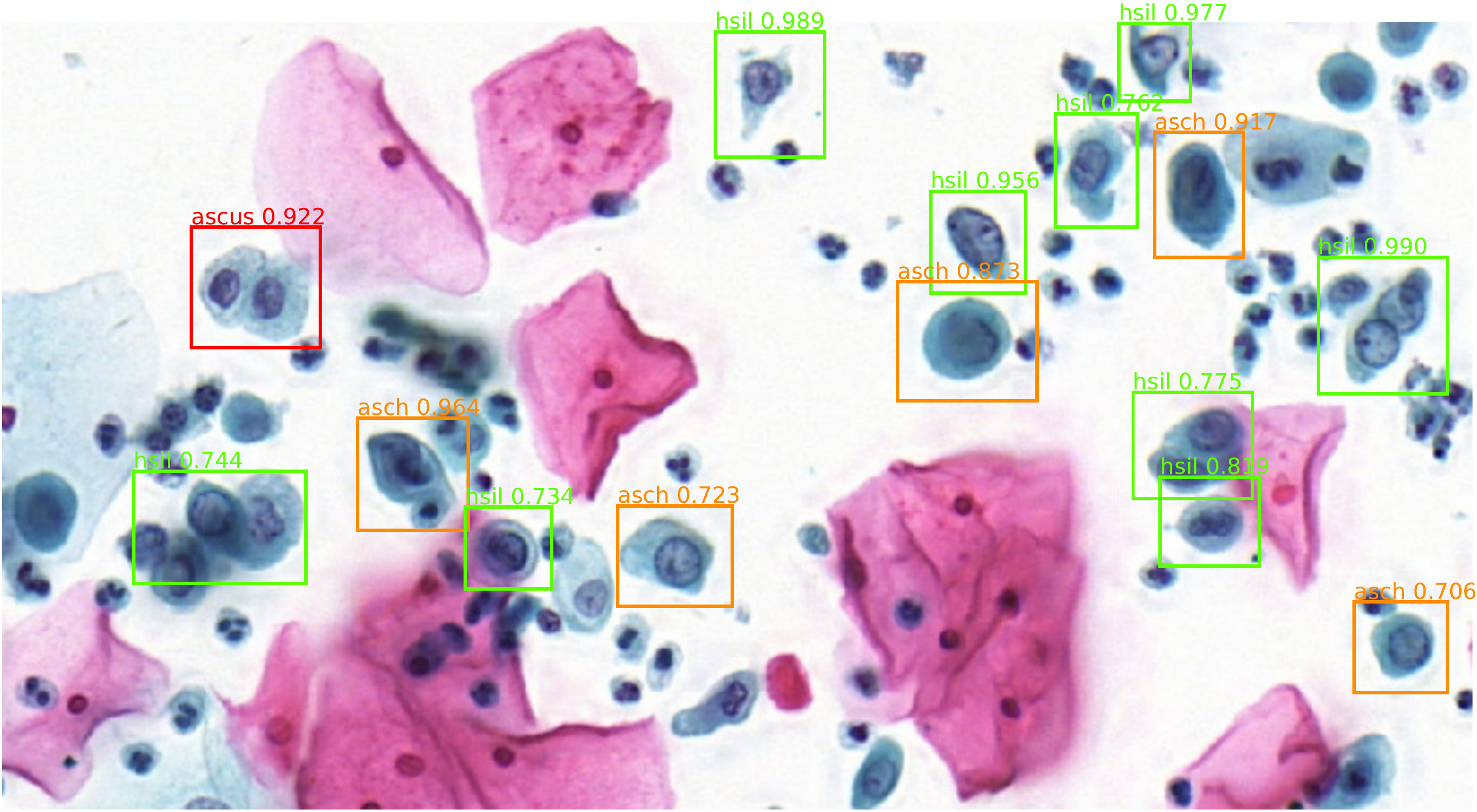}
\end{minipage}

\begin{minipage}{0.24\textwidth}
\includegraphics[height=3.2cm, width=4.3cm]{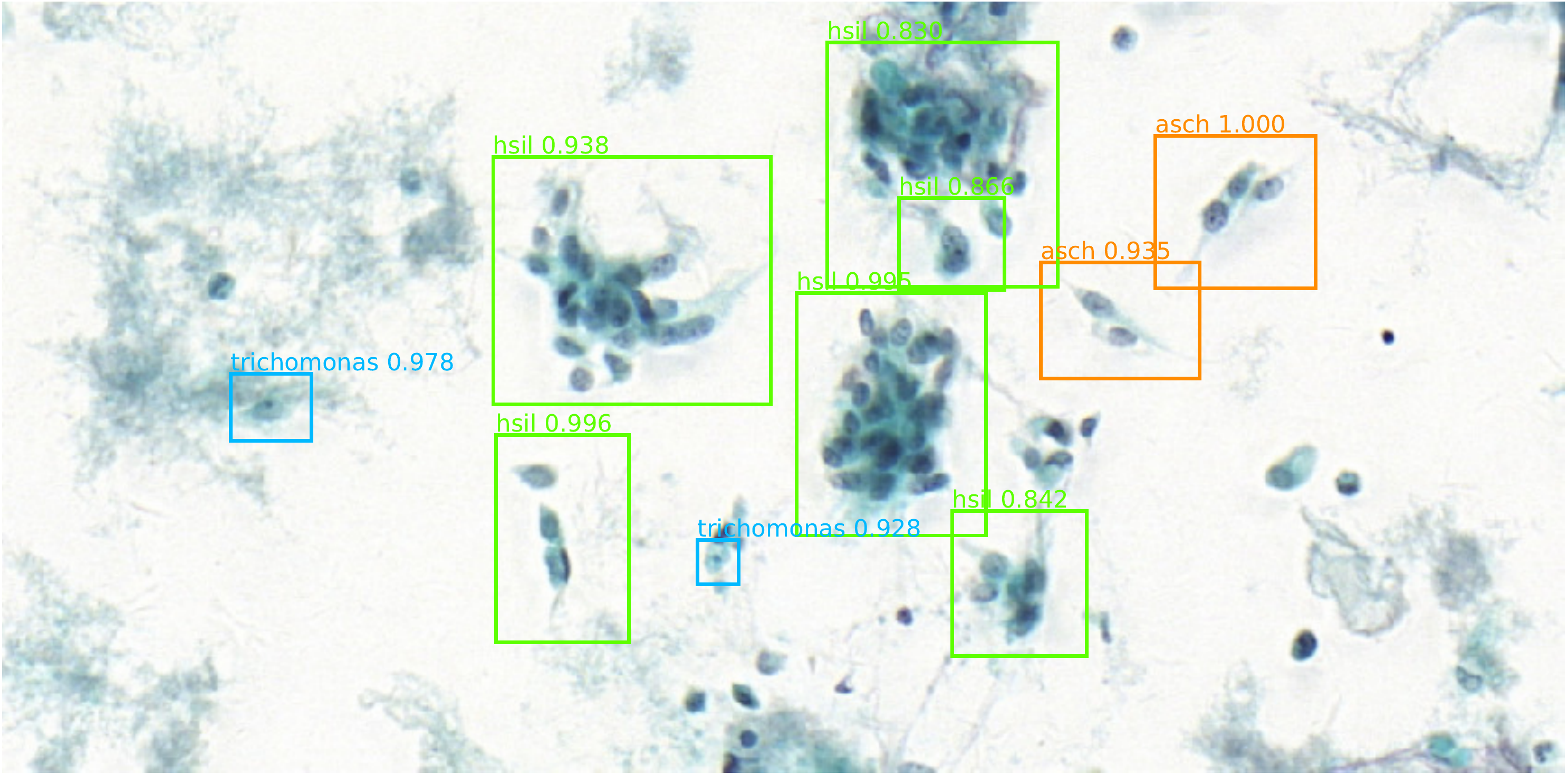}
\end{minipage}%
\begin{minipage}{0.24\textwidth}
\includegraphics[height=3.2cm, width=4.3cm]{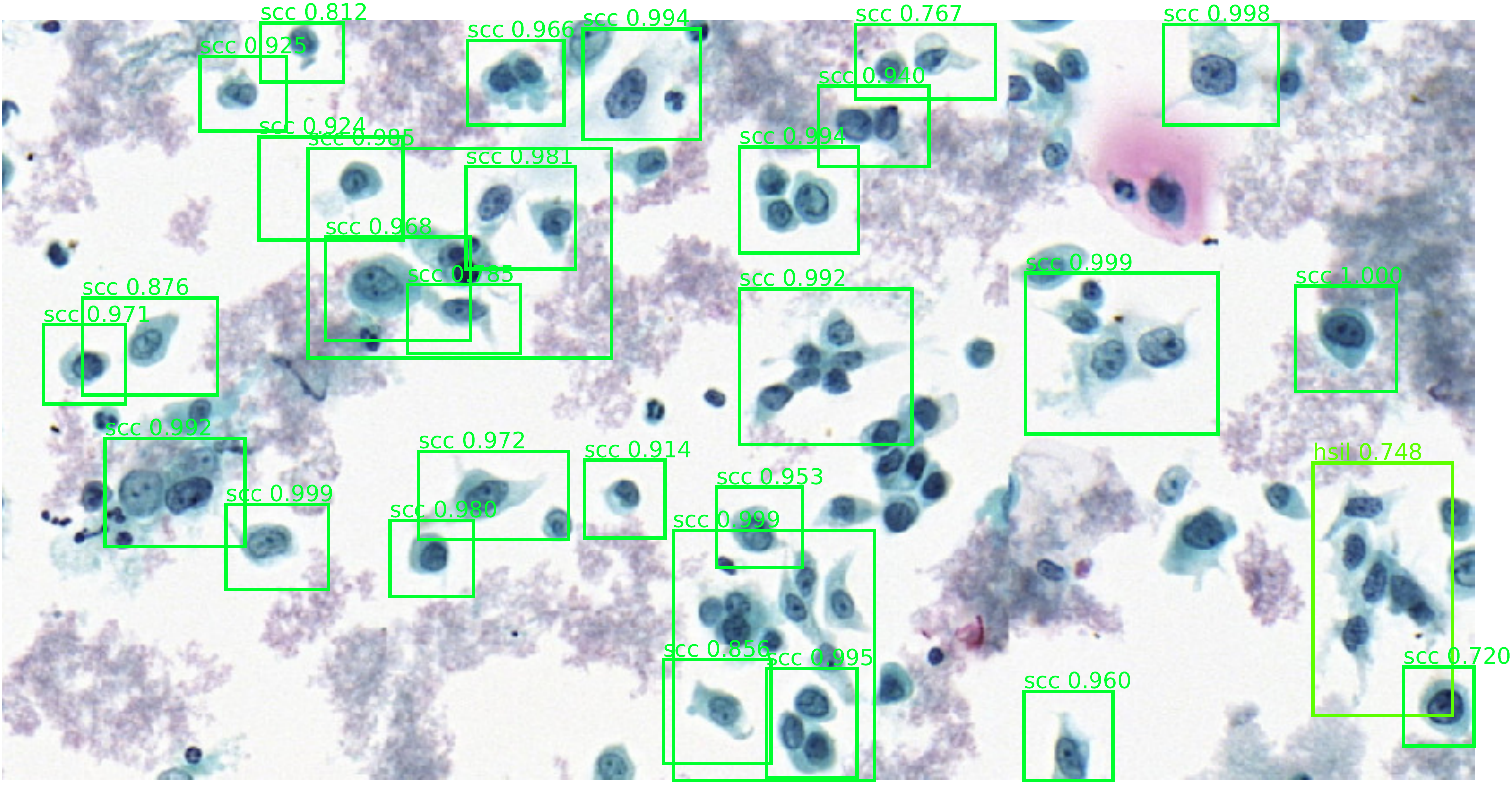}
\end{minipage}%
\begin{minipage}{0.24\textwidth}
\includegraphics[height=3.2cm, width=4.3cm]{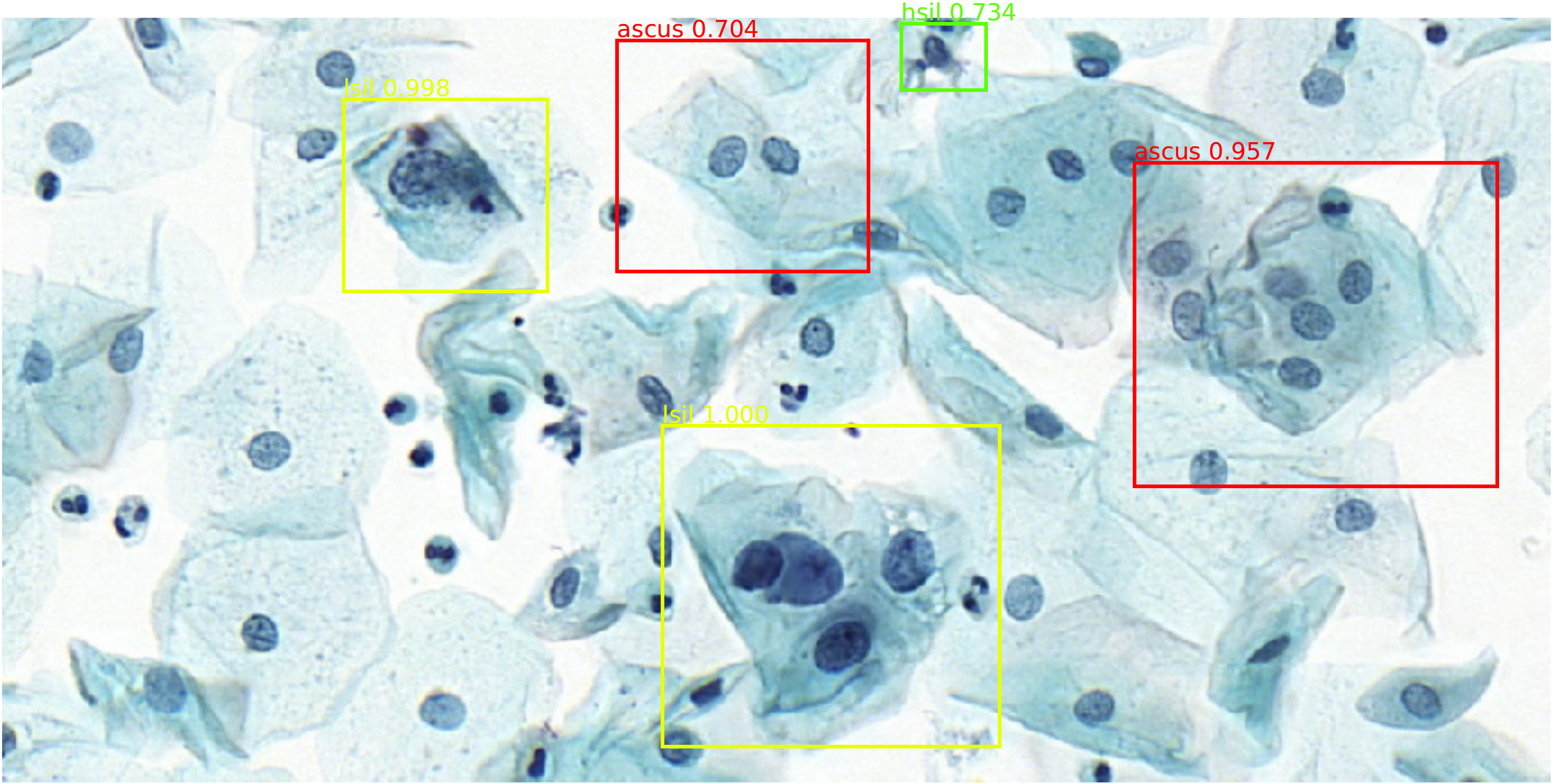}
\end{minipage}%
\begin{minipage}{0.24\textwidth}
\includegraphics[height=3.2cm, width=4.3cm]{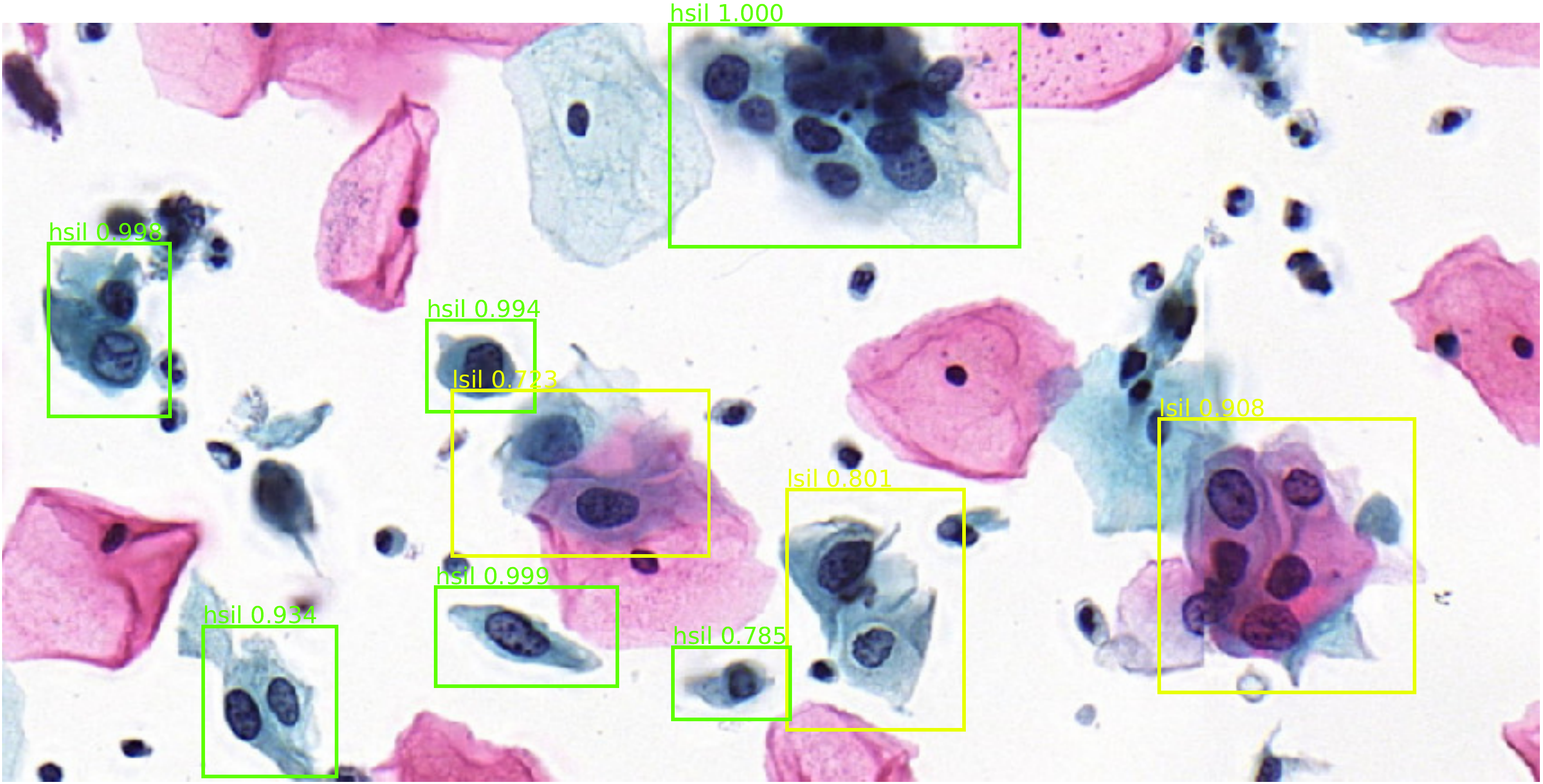}
\end{minipage}

\end{figure*}

For the cervical cell images, annotators are prone to take a higher threshold when 
label the objects due to the low discrimination of them. In addition, multiple nearby 
objects with the same category will be marked as one, so the performance of the model can not be well reflected by mAP \cite{girshick2014rich}. 
Therefore, the performance of the model is evaluated by using mAP and AR as a complement on test set. 
If the mAP does not decrease and the AR improves, it surely signifies the performance of model is improved.
 Herein, the results are reported in both mAP and AR. A summary of results can be found in Table \ref{tab:1} 
 and some detection results on the test set are shown in Fig.\ref{fig:7}.

\subsection{Learning the prototype representation of background}
In order to compare the effect of learning background, we randomly select some background samples from the proposals to obtain the features of background(model \texttt{A}).  As shown in Table \ref{tab:1},　the result is 31.4\% and the model \texttt{B} which is the model of learning the  prototype representation of background category has a mAP of  34.1\%. It indicates that the proposed method is better than random selection. It should be noted that because the prototype representation of the background category learns from the prototype representation of other categories, the gradient propagation will also have some effect on the optimization of other prototype representation. In order to make sure whether this effects is beneficial, we stop gradient propagation at the fork position in Fig. \ref{fig:2}(b). The performance of the model has declined with a mAP of 33.0\% and an AR of 52.6\%.

\subsection{Prototype representations of categories}
In our approach, as shown in Eq. \ref{equ:8}, we use all pyramid features to generate prototype representation of categories. Another choice is to only use the last level pyramid features as the category of prototype, i.e. $S (\{F_k^l\}) = F^5_k$. The model \texttt{B} makes use of all pyramid features to learn the protopype representations, however the model \texttt{C} only use the last level pyramid feature to learn the protopype representations. As shown in Table \ref{tab:1}, the result of model \texttt{B} is 34.1\% which is better than model \texttt{C} (32.7\%). The difference between models \texttt{D} and \texttt{E} is whether to use all pyramid features. The model \texttt{D} which has an mAP of 41.0\% is superior model \texttt{E} (38.9\%). They show that using all pyramid features performs is best. Because it can combine features of multiple scales, which not only have rich semantics but also take into account objects of different size.

\subsection{Head model}\label{section:4.3}
As mentioned before, in independent module, 
the box regression function $b(\cdot,\cdot)$ is the same as baseline model because experiment found 
that removing one layer will make the result worse. The model \texttt{B} is independent module and the model model \texttt{D} is shared module. The result lists in Table \ref{tab:1}. The model \texttt{B} has a mAP of 34.1\% and model \texttt{D} is 41.0\%. The results show shared module
performs much better than independent module. Furthermore, we drop the operation of bounding box regression in the head (model \texttt{F}). It's weird that it has a result 37.7\% which is better than model \texttt{B}. This phenomenon goes against common sense that 
twice bounding box regression are often better than just once. We empirically conjecture that the importance of classification should 
be more important than bounding box regression in our model \cite{liang2018object}. So we adjust the weight coefficient $\lambda$ in the Eq. \ref{equ:7}
 to balance the classification loss and bounding box regression loss. Here we select $\lambda=5$. The results are shown in Table \ref{tab:2}. After balancing loss, the performance of the model has been greatly improved. It confirms our guess.
 Moreover, by analyzing model \texttt{B} and model \texttt{D}, we find that the difference between them is not only 
 classification and bbox regression is independent, but also the comparison classifier of model \texttt{D} 
 is parameterized. After converting the comparison classifier of model \texttt{B} into parameterized(Fig. \ref{fig:3} (c)), 
 the result shows that it is better than model \texttt{D}.

\begin{table}[t]
\caption{Different way of selecting the reference samples.}
\label{tab:5}
\centering
\begin{tabular}{|c|c|c|c|}
\hline
\bfseries method & \bfseries fixed mode & \bfseries random mode & \bfseries t-SNE\\
\hline\hline
mAP &44.5  &42.8&\textbf{45.9} \\
\hline
AR   & 61.6  &61.0& \textbf{63.5}\\
\hline
\end{tabular}
\end{table}

\begin{table*}[!t]
\renewcommand{\arraystretch}{1.3}
\caption{The results of learning on different size datasets. We regard Faster-RCNN with Feature Pyramid Network as baseline model. The performance of Comparison detector and baseline model with training on different datasets are shown in the following:}
\label{tab:6}
\centering
\footnotesize
\begin{tabular}{|c|c|c|c|c|c|c|c|c|c|c|c|c|c|c|}
\hline
\bfseries method & \bfseries dataset & \bfseries AR & \bfseries mAP & \bfseries ascu &\bfseries asch &\bfseries lsil &\bfseries hsil &\bfseries scc & \bfseries agc & \bfseries trich & \bfseries cand & \bfseries flora & \bfseries herps&\bfseries actin\\
\hline\hline
\tabincell{c}{baseline\\model} & $D_s$ &12.9&6.6&11.0&2.0&23.7&21.6&0.0&3.5&0.0&11.5&0.0&0.0&0.0 \\
\hline
\tabincell{c}{Comparison\\detector} &$D_s$ &\textbf{35.7}&\textbf{26.3}&10.5&1.7&42.8&32.3&0.8&40.5&37.5&24.1&6.9&45.0&46.6\\
\hline
\tabincell{c}{baseline\\model} &$D_f$ &58.9&45.2&27.2&6.7&41.7&35.3&18.6&57.3&46.7&72.2&57.3&83.0&51.4\\
\hline
\tabincell{c}{Comparison\\detector} &$D_f$ &\textbf{63.5}&\textbf{45.9}&27.4&6.7&41.7&40.1&21.8&54.5&45.0&65.5&63.5&68.1&70.5\\
\hline
\end{tabular}
\end{table*}

\subsection{Optimizing comparison classifier}\label{sect:4.5}
We evaluate three distance metrics in the comparison classifier. The first is $\ell_2$-distance which means $d(P_m, F_k) = M(|F_k-P_m |^2) $. $M(\cdot)$ represents averaging function for tensor. The second is the parameterized $\ell_2$-distance, such as $d(P_m, F_k) = Conv_7(|F_k - P_m|^2)$. Similar to \cite{yang2018learning}, we also try to make the model to learn the metric function instead of the predefined ones. According to the result of Table \ref{tab:3}, parameterized $\ell_2$-distance shows the best performance which has an mAP of 44.5\%. So we ultimately adopt it. When $\lambda =5$, the result is shown in brackets. Combining with the results shown in Table \ref{tab:2},  we find that it is universal that the balance trick can improve performance in our model. So we adopt this trick in all the next experiments.

\subsection{Strategies for selecting reference samples}
We first evaluate the scheme of randomly choosing reference samples which includes two methods. 
The first is to randomly choose 3 instances of each category (this number is limited by GPU's memory)
 as the reference samples (\texttt{fixed mode}). The second one is to randomly select 5 candidates of each category in those objects. Then the model randomly selected three of the five candidates as templates during training, but five in testing (\texttt{random mode}). In addition, we also choose reference samples by applying t-SNE and K-means. During the training of t-SNE, we adopt the following parameters setting, i.e. the hyper-parameters are 30 for perplexity, 1 for learning rate, and 10 for label supervision. Throught t-SNE method, we can get the number of clusters for each category empirically, which as the \texttt{n\_clusters} parameter for K-means. The number of clusters are shown in Table \ref{tab:4}. Finally, according to the results of K-means, we choose the instance objects which is close to the center of clusters as reference samples. As show in Table \ref{tab:5}, the result of t-SNE and K-means is 45.9\%, however \texttt{fixed mode} is 44.5\% and \texttt{random mode} is 42.8\%. It indicates that the selection reference samples via t-SNE and K-means perform the best.

\subsection{Performance on training dataset $D_f$ and $D_s$}
As shown in Table \ref{tab:6}, Comparison detector has \textbf{0.7} mAP performance improvements when training on the $D_f$ dataset, and improves the AR by 4.6 points. Due to the special annotating situation as described in Section \ref{sect:4.1}, some correct predictions may be identified as false positives. Therefore, there is a significant increase in AR, but little improvement in mAP. When training on the $D_s$, Comparison detector is completely superior to baseline model. It achieves a state-of-the-art result on the test set with a mAP of 26.3\% compared to 6.6\%, which indicates our method alleviates the over fitting problem to some extent. Prototype representation in this model is generated by reference samples, however, it can be generated by other way, such as external memory. In the future work, we expect a better solution for the generation of prototype representations.

\section{Conclusion}
In this work, we propose to apply contemporary CNN-based object detection methods for automated cervical
cancer detection. To deal with the limited training dataset, we develop the comparison classifier into the state-of-the-art two-stage object detection method based on the comparison with the
reference samples of each category. Instead of manually choosing the reference samples of the background by some heuristic rules, 
we present a scheme to learn them from the data directly. We also investigate several important modules including the generation of prototype representations of each category and the design of head model for cervical cell/clumps detection. Experimental results show that compared with the baseline, our method improves the mAP by \textbf{19.7} points and the AR by \textbf{22.8} when trained on the small training dataset, and achieves better mAP and improves the AR by \textbf{4.6} when trained on the medium training dataset. It should be noticed that our algorithm directly operates on the whole image rather than the extracted patches based on the nuclei and hereby only need one forward propagation for each image, making the inference extremely efficient. In addition, the proposed method is \emph{flexible} to be intergraded into other proposal-based methods.

\section*{Acknowledgements}
This research was partially supported by the National Natural Science Foundation of China under Grant No. 61602522, the Natural Science Foundation of Hunan Province, China under Grant No.14JJ2008 and the Fundamental Research Funds of the Central Universities of Central South University under Grant No. 2018zzts577.
\clearpage
\bibliography{paper}
\end{document}